\documentclass[conference]{IEEEtran}
\IEEEoverridecommandlockouts
\usepackage{cite}
\usepackage{amsmath,amssymb,amsfonts}
\usepackage{graphicx}
\usepackage{textcomp}
\usepackage{xcolor}
\usepackage{cite}
\usepackage{amsmath,amssymb,amsfonts}
\usepackage{mleftright}
\usepackage{booktabs}         
\usepackage{algorithm}
\usepackage[noend]{algpseudocode}
\usepackage{threeparttable}
\usepackage{graphicx}
\usepackage{textcomp}
\def\BibTeX{{\rm B\kern-.05em{\sc i\kern-.025em b}\kern-.08em
    T\kern-.1667em\lower.7ex\hbox{E}\kern-.125emX}}
\begin{document}

\title{Online Decentralized  Federated Multi-task Learning With Trustworthiness in Cyber-Physical Systems
}

\author{\IEEEauthorblockN{Olusola  Odeyomi}
\IEEEauthorblockA{\textit{Department of Computer Science} \\
\textit{North Carolina Agricultural and Technical State University}\\
Greensboro, USA \\
otodeyomi@ncat.edu}
\and
\IEEEauthorblockN{Sofiat Olaosebikan}
\IEEEauthorblockA{\textit{School of Computing} \\
\textit{University of Glasgow}\\
Glasgow, UK \\
sofiat.olaosebikan@glasgow.ac.uk}
\and
\IEEEauthorblockN{Ajibuwa Opeyemi}
\IEEEauthorblockA{\textit{Department of Computer Science} \\
\textit{North Carolina Agricultural and Technical State University}\\
Greensboro, USA \\
oeajibuwa@aggies.ncat.edu}
\and
\IEEEauthorblockN{Oluwadoyinsola Ige}
\IEEEauthorblockA{\textit{Department of Computer Science} \\
\textit{North Carolina Agricultural and Technical State University}\\
Greensboro, USA \\
orige@aggies.ncat.edu}
}

\maketitle

\begin{abstract}
Multi-task learning is an effective way to address the challenge of model personalization caused by high data heterogeneity in federated learning. However, extending multi-task learning to the online decentralized federated learning setting is yet to be explored. The online decentralized federated learning setting considers many real-world applications of federated learning, such as autonomous systems, where clients communicate peer-to-peer and the data distribution of each client is time-varying. A more serious problem in real-world applications of federated learning is the presence of Byzantine clients. Byzantine-resilient approaches used in federated learning work only when the number of Byzantine clients is less than one-half the total number of clients. Yet, it is difficult to put a limit on the number of Byzantine clients within a system in reality. 
However, recent work in robotics shows that it is possible to exploit cyber-physical properties of a system to predict clients' behavior and assign a trust probability to received signals. This can help to achieve resiliency in the presence of a dominating number of Byzantine clients. Therefore, in this paper, we develop an online decentralized federated multi-task learning algorithm to provide model personalization and resiliency when the number of Byzantine clients dominates the number of honest clients. Our proposed algorithm leverages cyber-physical properties, such as the received signal strength in wireless systems or side information,  to assign a trust probability to local models received from neighbors in each iteration. Our simulation results show that the proposed algorithm performs close to a Byzantine-free setting.
\end{abstract}

\begin{IEEEkeywords}
Byzantine attacks, federated learning, multi-task learning, online learning, regret
\end{IEEEkeywords}

\section{Introduction}
Federated learning is a distributed machine learning paradigm that allows multiple clients to train their data collaboratively without sharing it \cite{mcmahan2017communication}. Federated learning has wide applications, such as in wireless systems, autonomous systems \cite{zeng2022federated}, computer vision \cite{shenaj2023federated}, large language models \cite{gupta2022recovering}, etc. Conventional federated learning architecture consists of a central server communicating back and forth with clients. Recent architecture favors a fully decentralized setting where clients communicate directly without the need to route through a central server \cite{beltran2023decentralized,lalitha2018fully}. This mitigates against a single point of failure attack and congestion at the central server.

Federated learning typically assumes that clients' data distributions remain constant over time. However, this assumption reduces its applicability in many real-world scenarios where data distributions change dynamically \cite{rizk2020dynamic,fan2023uav,wu2024flexible}. For example, in wireless settings, the environment constantly evolves, leading to time-varying data collected from observing environmental phenomena.
Thus, it is important to design federated learning algorithms that can work with time-varying data distributions.

The goal of federated learning is to train a global model that performs well across different clients' test datasets. However, due to the non-independent and non-identically distributed (non-IID) nature of client data, a single global model may not generalize effectively across heterogeneous test datasets, leading to unfair performance for some clients. To address this issue, model personalization techniques such as clustering \cite{briggs2020federated}, meta-learning \cite{fallah2020personalized}, and multi-task learning \cite{smith2017federated, marfoq2021federated, corinzia2019variational} have been proposed. Clustering involves training multiple global models and assigning clients to the model that best fits their data. Clients with similar data distributions are grouped around the same model. However, this approach is computationally expensive, as it requires training multiple global models. Meta-learning trains a global model to serve as a meta-initializer, which is further fine-tuned using each client’s dataset. However, the Hessian computation involved in meta-learning can be computationally intensive. While ignoring the Hessian reduces complexity, it also weakens the generalization ability of the meta-initializer.
Multi-task learning is another effective approach for personalization in federated learning. It personalizes models by either penalizing the global cost function or introducing constraints to it. Multi-task learning is a popular method for model personalization in federated learning.

Federated learning algorithms, particularly in wireless applications, are vulnerable to Byzantine attacks \cite{shi2022challenges,fang2020local}. In these attacks, malicious clients send arbitrary local model updates either to the server in conventional federated learning or to neighboring clients in fully decentralized settings, deliberately steering model convergence away from the global optimum. A Byzantine client can simultaneously send different manipulated updates to multiple clients, and research has shown that even a single Byzantine client can disrupt convergence, regardless of the total number of participating clients \cite{odeyomi2023online}. One effective mitigation strategy is to apply statistical analysis to the received model updates obtained from training data, using techniques such as computing the mean or median, or filtering out outliers. Methods based on statistical analysis include geometric median \cite{cohen2016geometric}, coordinate-wise trimming \cite{yin2018byzantine}, BRIDGE \cite{fang2022bridge}, UBAR \cite{guo2021byzantine}, and Brydie \cite{yang2019byrdie}. However, a major limitation of this approach is that it only works when the number of Byzantine clients is less than one-half of the total number of clients. If Byzantine clients outnumber honest clients, these statistical techniques fail. Currently, no federated learning algorithm has been developed that can effectively mitigate Byzantine attacks when malicious clients dominate the network.

Therefore, in this paper, we address a combination of all these challenges. We design an online decentralized federated multi-task learning algorithm that is robust against Byzantine attacks, where the number of Byzantine clients dominates the federated learning. We model the learning as a constrained optimization problem which is solved using the regularized Lagrangian optimization approach. To mitigate Byzantine attacks, we leverage the cyber-physical properties guiding clients' communication. Existing research has shown that in cyber-physical systems, physical channels, rather than data, can be exploited to predict client trustworthiness \cite{yemini2022resilience, yemini2025resilient}. For instance, in wireless and multi-robot systems, the received signal strength can be analyzed to determine if a received packet is malicious. A confidence score,  known as the trust probability $\alpha\in[0,1]$, is then assigned to the received signal and utilized to predict malicious signals even when malicious clients dominate the systems \cite{gil2017guaranteeing}. Thus, this paper utilizes a trust probability to mitigate Byzantine attacks in online decentralized federated multitask learning for the first time. Our simulation results show that the proposed algorithm can achieve sublinear regret and constraint violation in the presence of a dominating number of Byzantine clients with close performance to the scenario where there are no Byzantine clients.

\section{Problem Formulation}
Consider $V$ clients communicating peer-to-peer and forming a directed graph network $\mathcal{G}=(\mathcal{V},\mathcal{E})$ with $\mathcal{E}\subseteq \mathcal{V}\times\mathcal{V}$ representing the edge set. Let $\mathcal{N}_v=\{u\in\mathcal{V}|(u,v)\in\mathcal{E}\}$ denote the neighborhood of client $v\in\mathcal{V}$. Each client $v\in\mathcal{V}$ samples a minibatch dataset $D_{v,t}$ from its time-varying data distribution $\mathcal{D}_{v,t}$ at time $t$ and updates its local model with the minibatch dataset. In every iteration $t$, each client $v$ decides its local model $\vec{x}_{v,t}\in\mathcal{X}\subseteq\mathbb{R}^d$ with the time-varying loss function $f_{v,t}:\mathcal{X}\rightarrow \mathbb{R}$. The local models of any two connected clients are subject to a constraint $g_{uv}:\mathcal{X}\times \mathcal{X}\rightarrow \mathbb{R}$, i.e., $g_{uv}(\vec{x}_{u,t},\vec{x}_{v,t})$ $\forall (u,v)\in\mathcal{E}$. The constraint function is symmetric, i.e., $g_{uv}(\vec{x}_{u,t},\vec{x}_{v,t})=g_{vu}(\vec{x}_{v,t},\vec{x}_{u,t})$. Moreover, the constraint function $g_{uv}$ is assumed to be fixed for all clients in all iterations and available at the start of the collaborative training. An example of such a constraint is the quadratic constraint that models proximity or similarities among the clients' local models \cite{bhattarai2016cp,koppel2017proximity,bedi2019asynchronous}. The goal of federated learning is to minimize the training cost over the time horizon where each client has a specific task that has some relationship to other tasks. This is given as
\begin{equation}
\begin{split}
\min_{\vec{x}\in\mathcal{X},\forall v} \mathbb{E}_{J\in \Gamma}\bigg[\sum_{t=1}^T \sum_{v=1}^V f_{v,t}(\vec{x}_{v,t};D_{v,t})\bigg] \\
\text{subject to}\\
 \mathbb{E}_{J\in \Gamma}\bigg[ \sum_{t=1}^T g_{uv}(\vec{x}_{u,t},\vec{x}_{v,t})\bigg]\leq 0 \quad \forall (u,v)\in\mathcal{E}
\end{split}
\label{eqn 1}
\end{equation}
where $J= \{D_{v,t}\}_{1\leq v\leq V;1\leq t\leq T}$ and $\Gamma = \{\mathcal{D}_{v,t}\}_{1\leq v\leq V;1\leq t\leq T}$. The expectation $\mathbb{E}[\quad \cdot \quad]$ accounts for the randomness of $D_{v,t}$. The data distribution across the clients is both time-varying and heterogeneous.
It is impossible to solve (\ref{eqn 1}) online because it requires the knowledge of future loss functions. However, we can apply sequential decision learning to solve the multi-task learning problem. The performance metric is the static regret that measures the performance of an online algorithm with an oracle with hindsight knowledge of the optimal task-specific models. This is given as
\begin{equation}
\begin{split}
   Reg(T) = \\\mathbb{E}_{J\in\Gamma}\bigg[\sum_{t=1}^T \sum_{v=1}^V f_{v,t}(\vec{x}_{v,t};D_{v,t}) - \sum_{t=1}^T \sum_{v=1}^V f_{v.t}(\vec{x}_v^*;D_{v,t})\bigg]  
   \label{eqn 2}
   \end{split}
\end{equation}
where $\vec{x}^* = [(\vec{x}^*_1)^\mathbb{T},...,(\vec{x}^*_V)^\mathbb{T})]^\mathbb{T}$ is the collection of the optimal task-specific models across the clients, and it is defined as
\begin{equation}
\begin{split}
(\vec{x}^*_1,...,\vec{x}^*_V)\in\arg\min_{\vec{x}_v\in\mathcal{X},\forall v}   \mathbb{E}_{J\in\Gamma}\bigg[\sum_{t=1}^T\sum_{v=1}^V f_{v,t}(\vec{x}_v;D_{v,t})\bigg]\\
\text{subject to}\quad g_{uv}(\vec{x}_{u,t},\vec{x}_{v,t})\leq 0 \quad\forall (u,v)\in\mathcal{E}
\end{split}
\label{eqn 3}
\end{equation}
It should be observed that the optimal task-specific models satisfy a per-round constraint in (\ref{eqn 3}), rather than the long-term constraint in (\ref{eqn 1}). This is because according to Proposition 4 of \cite{mannor2009online} and Proposition 2.1 of \cite{sun2017safety}, no online algorithm can obtain a sublinear regret if it competes with the best optimal task-specific models that satisfy (\ref{eqn 1}) rather than (\ref{eqn 3}). Another metric is constraint violation defined as
\begin{equation}
    Vio_{uv}(T) = \mathbb{E}_{J\in\Gamma}\bigg[\sum_{t=1}^T g_{uv}(\vec{x}_{u,t},\vec{x}_{v,t})\bigg] \quad\forall (u,v)\in\mathcal{E}.
\end{equation}

It should be observed that (\ref{eqn 1}) and (\ref{eqn 3}) assume that all the clients are honest. However, in practical federated learning, some clients can be Byzantine. Byzantine clients act arbitrarily and are difficult to identify. Therefore, honest clients are saddled with the responsibility of learning which clients to trust. Hence, we reformulate the regret definition for honest client $v$ in the presence of $b =|\mathcal{V}_b|$ Byzantine clients as follows:
\begin{equation}
\begin{split}
   Reg(T) = \\\mathbb{E}_{J\in\Gamma}\bigg[\sum_{t=1}^T \sum_{v=1}^{V-b} f_{v,t}(\vec{x}_{v,t};D_{v,t}) - \sum_{t=1}^T \sum_{v=1}^{V-b} f_{v.t}(\vec{x}_v^*;D_{v,t})\bigg]  
   \label{eqn 2}
   \end{split}
\end{equation}
Similarly, the constraint violation can be redefined as 
\begin{equation}
    Vio_{uv}(T) = \mathbb{E}_{J\in\Gamma}\bigg[\sum_{t=1}^T g_{uv}(\vec{x}_{u,t},\vec{x}_{v,t})\bigg] \quad\forall (u,v)\in\mathcal{E}/\mathcal{K}.
\end{equation}
where $\mathcal{K}:=\{(k,v) \& (v,k)|k \in \mathcal{V}_b\subset\mathcal{V}; \mathcal{V}_b = \{1,...,b\}; v\in\mathcal{V}_h=\mathcal{V}/\mathcal{V}_b\}$. The optimal task-specific models for the honest clients are given as 
\begin{equation}
\begin{split}
(\vec{x}^*_1,...,\vec{x}^*_{V-b})\in\\
\arg\min_{\vec{x}_v\in\mathcal{X},\forall v\in\mathcal{V}/\mathcal{V}_b}   \mathbb{E}_{J\in\Gamma}\bigg[\sum_{t=1}^T\sum_{v=1}^{V-b} f_{v,t}(\vec{x}_{v};D_{v,t})\bigg]\\
\text{subject to}\quad g_{uv}(\vec{x}_{u,t},\vec{x}_{v,t})\leq 0 \quad\forall (u,v)\in\mathcal{E}/\mathcal{K}.
\end{split}
\label{eqn 4}
\end{equation}
The sublinearity of the regret which determines the goodness of the online algorithm is defined as $Reg(T)\leq o(T)$. Similarly, the sublinearity of the constraint violation is defined as $Vio_{uv}(T)\leq o(T)\quad\forall (u,v)\in\mathcal{E}$. The time-average regret is given as $\frac{Reg(T)}{T}\leq o(1)$ and the time-average constraint violation is given as $\frac{Vio_{uv}(T)}{T}\leq o(1)\quad\forall (u,v)\in\mathcal{E}$. Both $Reg(T)$ and $Vio_{uv}(T)$ are asymptotically non-positive as $T\rightarrow \infty$. Hence, the asymptotic performance of $\{\vec{x}_{v,t}\}_{1\leq v\leq V; 1\leq t\leq T}$ with $T\rightarrow \infty$ is no worse than the optimal task-specific models $\vec{x}^*$ in a time-average sense.

\section{Algorithm Formulation}
This section aims to formulate a saddle-point algorithm for training honest clients in decentralized federated multi-task learning settings. We start with formulating the regularized Lagrangian optimization of (\ref{eqn 1}) while assuming that all clients are honest. Then, we extend the formulation to address the presence of Byzantine clients. The regularized Lagrangian optimization of (\ref{eqn 1}) at time $t$ in the offline setting is given as
\begin{equation*}
\begin{split}
 \mathcal{L}_t(\vec{x},\vec{\Lambda}) =  \\ \sum_{v=1}^V \bigg[f_{v,t}(\vec{x}_v;D_{v,t}) + \sum_{u\in\mathcal{N}_v} \bigg(\lambda_{vu}g_{vu}(\vec{x}_v,\vec{x}_u)- \frac{\delta\eta}{2}\lambda_{vu}^2\bigg)\bigg] 
 \end{split}
\end{equation*}
\begin{equation}
\begin{split}
   = f_t(\vec{x},J) + \langle {\Lambda},G(\vec{x}) \rangle -\frac{\delta \eta}{2}||{\Lambda}||^2 \quad \forall t\in[T]
   \label{eqn 8}
   \end{split}
\end{equation}
where $\vec{x} = [\vec{x}_1^\mathbb{T},...,\vec{x}_V^\mathbb{T}]^\mathbb{T}$ is the primal variable; $f_t(\vec{x},J) = \sum_{v=1}^V f_{v,t}(\vec{x}_v; D_{v,t})$ is the global loss function; $\lambda_{vu}>0$ is the dual variable associated with the constraint $g_{vu}(\cdot)\leq 0 \quad \forall (v,u)\in\mathcal{E}$; $\Lambda_v \in \mathbb{R}_+^{\mathcal{N}_v}$; $\Lambda = [\Lambda_1^\mathbb{T},...,\Lambda_V^\mathbb{T}]$; and $G(\vec{x})$ is the concatenation of all constraints.  The regularizer $-\frac{\delta \eta}{2}||\Lambda||^2$ suppresses the growth of the dual parameter to ensure the stability of the resulting algorithm. The stepsize is $\eta>0$ and the control parameter is $\delta>0$. The saddle-point of (\ref{eqn 8}) can be obtained by alternating gradient updates on both the primal and dual variables. The gradients of (\ref{eqn 8}) with respect to both the primal and dual variables are
\begin{equation}
\begin{split}
  \triangledown_{\vec{x}_v}\mathcal{L}_t(\vec{x},\Lambda) =   \triangledown_{\vec{x}_v}f_{v,t}(\vec{x}_v;D_{v,t}) + \\ \sum_{u\in\mathcal{N}_v}[\lambda_{vu}\triangledown_{\vec{x}_v}g_{vu}(\vec{x}_v,\vec{x}_u)+ \lambda_{uv}\triangledown_{\vec{x}_v}g_{uv}(\vec{x}_u, \vec{x}_v)], \quad \forall v\in\mathcal{V}
  \end{split}
  \label{eqn 9}
\end{equation}

\begin{equation}
 \frac{\partial}{\partial \lambda_{vu}} \mathcal{L}_t(\vec{x},\Lambda) = g_{vu}(\vec{x}_v,\vec{x}_u)-\delta\eta\lambda_{vu}, \quad\forall (v,u)\in\mathcal{E}.
 \label{eqn 10}
\end{equation}
Computing $\triangledown_{\vec{x}_v}\mathcal{L}_t(\vec{x},\Lambda)$ and $\frac{\partial}{\partial\lambda_{vu}}\mathcal{L}_t(\vec{x},\Lambda)$ require the knowledge of the local models of all neighboring clients $u\in\mathcal{N}_v$.  However, some of the neighboring clients may be Byzantine. Therefore, (\ref{eqn 9}) and (\ref{eqn 10}) must be modified to handle Byzantine infiltration in the algorithm. 

\subsection{Finding the Set of Trusted Neighbors}
We employ a stochastic probability of trust $\alpha_{vu}$ to determine the set of trusted neighbors, which is defined as follows:

\textbf{Definition 1:} For every honest client $v$ and neighborhood $\mathcal{N}_v$, the random variable $\alpha_{vu}(t)\in [0,1]$ represents the probability that client $u$ is a trusted neighbor of client $v$ at time $t$. 

Some existing works are focused on deriving the probabilistic trust values, such as \cite{gil2017guaranteeing}. This is not the focus here, rather it is to develop a federated multi-task learning algorithm that supports resilient convergence in the presence of Byzantine agents with the assumption that $\alpha_{vu}(t)$ is known from the physicality of any cyber-physical systems. Such an assumption is common in existing works \cite{yemini2022resilience,yemini2025resilient} where the focus is on algorithm development that guarantees resiliency regardless of how the trust values are derived. Thus, it is assumed that each honest client $v$ can observe $\alpha_{vu}(t)$ for $u\in\mathcal{N}_v$ which corresponds to the legitimacy of the received information. When $\alpha_{vu}(t)\geq0.5$, this indicates legitimate transmission, and when $\alpha_{vu}(t)<0.5$, this indicates illegitimate transmission in a probabilistic sense at time $t$. However, misclassification of transmission may occur during the early stage of training the algorithm because the algorithm has not learned enough, but it exponentially decreases with time. 

\textbf{Assumption 1 \cite{yemini2025resilient}:} (i) [Sufficiently Connected Graph] The subgraph formed by honest clients is connected. (ii) [Homogeneity of Trust Values] There are scalars $E_v\geq0$ and $E_b< 0$, such that 
\begin{equation}
\begin{split}
E_v := \mathbb{E}[\alpha_{vu}(t)]-0.5 \quad \forall v\in \mathcal{V}_h \quad \text{and}  \quad u\in\mathcal{N}_v\cap \mathcal{V}_h   \\
E_b := \mathbb{E}[\alpha_{vu}(t)] - 0.5  \quad \forall v\in \mathcal{V}_h \quad \text{and}  \quad u\in\mathcal{N}_v\cap \mathcal{V}_b.  
\end{split}
\end{equation}
where $\mathcal{V}_h$ and $\mathcal{V}_b$ are the set of honest and Byzantine clients respectively. 
(iii) [Independence of Trust Variables] The trust probability $\alpha_{vu}(t)$ is independent across all honest clients $v$ and neighboring set $\mathcal{N}_v$ at each time $t$. Moreover, the set of all trust probabilities $\{\alpha_{vu}(t)\}$ is independent and identically distributed.

Considering the sum of all trust probabilities up to time $t-1$ denoted as $\beta_{vu}(t)$, we have
\begin{equation}
 \beta_{vu}(k):=\sum_{k=0}^{t-1}(\alpha_{vu}(t)-0.5)\quad\text{for}\quad k\geq 1, v\in\mathcal{V}_h, u\in\mathcal{N}_v.  
 \label{eqn 12}
\end{equation}
where $\beta_{vu}(0)=0$. Following Definition 1, $\beta_{vu}(t)\geq0$ for a transmission from an honest neighbor and $\beta_{vu}< 0$ for a transmission from a Byzantine neighbor. The probability of misclassification decays over time according to Lemma 1.

\textbf{Lemma 1:} Consider the random variable $\beta_{vu}(t)$ defined in (\ref{eqn 12}). Then, for every $t\geq 0$, and every $v\in\mathcal{V}_h$, $u\in\mathcal{N}_v \cap \mathcal{V}_h$
\begin{equation}
  Pr(\beta_{vu}(t)< 0)  \leq \max\{\exp(-2tE^2_v),\mathbb{I}_{\{E_v<0\}}\}.
\end{equation}
Furthermore, for every $t\geq 0$, and every $v\in\mathcal{V}_h$, $u\in\mathcal{N}_v \cap \mathcal{V}_b$,
\begin{equation}
 Pr(\beta_{vu}(t)\geq 0)\leq \max\{\exp(-2tE_b^2),\mathbb{I}_{\{E_m\geq0\}}\}.   
\end{equation}

\textbf{Corollary 1:} A finite random time $T_f$ exists such that 
\begin{equation}
\begin{split}
 \beta_{vu}(t)\geq 0 \quad \forall t\geq T_f, v\in\mathcal{V}_h, u\in\mathcal{N}_v\cap\mathcal{V}_h,\\
 \beta_{vu}<0 \quad\forall t\geq T_f, v\in\mathcal{V}_h, u\in\mathcal{N}_v\cap\mathcal{V}_b
 \end{split}
\end{equation}
and there also exists 
\begin{equation}
\begin{split}
 \beta_{vu}(T_f-1)<0 \quad\text{for some}\quad u\in\mathcal{N}_v\cap\mathcal{V}_h,\quad\text{or}\\
 \beta_{vu}(T_f-1)\geq 0 \quad\text{for some}\quad j\in\mathcal{N}_v\cap\mathcal{V}_b
\end{split} 
\end{equation}
The set of trusted neighbors of client $v$ at time $t$ is defined as $\mathcal{N}_v^h(t):= \{u\in\mathcal{N}_v|\beta_{vu}(t)\geq 0\}$, and $\mathcal{E}_v^h(t)$ is the edge-set of all edges connecting honest client $v$ and every $u\in\mathcal{N}_v^h(t)$.

\textbf{Proof:} The Proof follows from Proposition 1 in \cite{yemini2021characterizing}.

\subsection{Resilient Saddle Point-Based Federated Multitask Learning}
We return to the gradients of regularized Lagrangian optimization with respect to both the primal and dual variables in (\ref{eqn 9}) and (\ref{eqn 10}) to handle Byzantine attacks from Byzantine clients. The resilient gradients for the online setting become
\begin{equation}
\begin{split}
  \vec{q}_{v,t} :=   \triangledown_{\vec{x}_v}f_{v,t}(\vec{x}_{v,t};D_{v,t}) + \\ \sum_{u\in\mathcal{N}_v^h(t)}[(\lambda_{vu,t}+\lambda_{uv,t})\triangledown_{\vec{x}_v}g_{vu}(\vec{x}_{v,t},\vec{x}_{u,t})], \quad \forall v\in\mathcal{V}_h
  \end{split}
  \label{eqn 17}
\end{equation}

\begin{equation}
 r_{vu,t} := g_{vu}(\vec{x}_{v,t},\vec{x}_{u,t})-\delta\eta\lambda_{vu,t}, \quad\forall (v,u)\in\mathcal{E}_v^h(t).
 \label{eqn 18}
\end{equation}

\begin{algorithm}
\caption{Online Federated Multitask Learning With Trustworthiness for Each Honest Client $v$}

\begin{algorithmic}[1]

\State \textbf{Input}: Time duration $T$, step sizes $\eta>0$, regularization $\delta>0$.
\State \textbf{Output:} $\vec{x}_{v,T}$.
\State Initialization: $\vec{x}_{v,1}$, $\lambda_{vu,1}$ 
\For{$t = 1, ... ,T$}
\State   Send $\vec{x}_{v,t}$ and $\lambda_{vu,t}$ to $\forall u\in\mathcal{N}_v$. 
\State Receive $\vec{x}_{u,t}$  and $\lambda_{uv,t}$ from $ \forall u\in\mathcal{N}_v$. 
\State Observe trust probability $\alpha_{vu}(t)\quad \forall u\in\mathcal{N}_v$ from inter-client interactions.
\State Compute $\beta_{vu}(t)$ using (\ref{eqn 12}).
\State Compute trusted set $\mathcal{N}_v^h(t)_=\{u\in\mathcal{N}_v|\beta_{vu}(t)\geq 0\}$.
\State Obtain the loss $f_{v,t}(\vec{x}_{v,t};D_{v,t})$ and constraint \indent $g_{v,u}(\vec{x}_{v,t},\vec{x}_{u,t})\quad \forall u\in\mathcal{N}_v^h(t)$.
\State Compute $\vec{q}_{v,t}$ using equation (\ref{eqn 17}).
\State  Compute ${r}_{vu,t}$ using equation (\ref{eqn 18}).
\State Compute primal variable $\vec{x}_{v,t+1}= \prod_{\mathcal{X}}(\vec{x}_{v,t}-\eta\vec{q}_{v,t})$.
\State Update dual variable $\lambda_{vu,t+1}=[\lambda_{vu,t}+\eta r_{vu,t}]_+ \quad\forall u\in\mathcal{N}_v^h(t)$, with $[\cdot]_+=\max(\cdot,0)$.
\EndFor
\end{algorithmic}
\end{algorithm}

From Algorithm 1, the local model and dual parameter of honest client $v$ is initialized in Step 3. From Steps 4 - 14, the algorithm iterates over $T$ rounds. Each iteration round $t$ starts from Step 5 and is illustrated as follows:  In Step 5, honest client $v$ sends its local model and dual parameter to its neighbors $u\in\mathcal{N}_v$ which may include Byzantine clients. In Step 6, honest client $v$ receives local models and dual parameters from all neighbors $u\in\mathcal{N}_v$, which may include Byzantine clients. In Step 7, honest client $v$ analyzes the inter-agent interactions signal such as the received signal strength in wireless communication, and assigns a trusted probability $\alpha\in[0,1]$ to the received local models from the neighbors $u\in\mathcal{N}_v$. In Step 8, honest agent $v$ computes the beta value $\beta_{vu}(t)$ using Equation (\ref{eqn 12}) and proceeds in Step 9 to compute a trusted set of neighbors $\mathcal{N}_v^h(t)$, which varies over time depending on the $\beta$ value. In Step 10, honest client $v$ trains its local model with its time-varying data drawn from the time-varying data distribution and obtains the loss $f_{v,t}(\vec{x}_{v,t},D_{v,t})$. Furthermore, it computes the constraint value $g_{vu}(\vec{x}_{v,t},\vec{x}_{u,t})$ that measures the similarity of the local model of honest client $v$ and trusted clients $u\in\mathcal{N}_v^h(t)$ in iteration $t$. Then, in Step 11, it computes the gradient of the Lagrangian function denoted as $\vec{q}_{v,t}$ with respect to the primal parameter $\vec{x}_v$ using equation (\ref{eqn 17}). Also, Step 12 computes the gradient of the Lagrangian function denoted as $r_{vu,t}$ with respect to the dual parameter using equation (\ref{eqn 18}). Step 13 uses projected gradient descent to update the local model  $\vec{x}_{v,t+1}$. Step 14 updates the dual parameter $\lambda_{vu,t+1}$ using gradient ascent with $[\cdot]_+ :=\max(\cdot,0)$. This process repeats for $T$ iterations for each honest client $v$. 

\section{Theoretical Results}
In this section, we present the theoretical results.

\textbf{Assumption 2:} The set $\mathcal{X}$ is convex, closed, and bounded. That is, there exist a $r>0$ such that $||\vec{x}||\leq r, \quad \forall \vec{x}\in\mathcal{X}$.

\textbf{Assumption 3:} The loss function $f_{v,t}(\cdot)$ is convex in $\vec{x}_v\in\mathcal{X}$ for all honest clients $v\in \mathcal{V}_h$. The similarity constraint $g_{vu}(\vec{x}_v,\vec{x}_u)$ is jointly convex in $\vec{x}_v$ and $\vec{x}_u$ for any $v\in\mathcal{V}_h$, $u\in\mathcal{N}_v^h(t)$, and $(v,u)\in\mathcal{E}_v^h(t)$.

\textbf{Assumption 4:} There exists a constant $G>0$ such that for any $\vec{x}_v$, $\vec{x}^\prime_v\in\mathcal{X}$, $v\in\mathcal{V}$, and $t\in\{1,...,T\}$, 
\begin{equation}
    |f_{v,t}(\vec{x}_v)-f_{v,t}(\vec{x}_v^\prime)|\leq G||\vec{x}_v-\vec{x}^\prime_v||
\end{equation}

\textbf{Assumption 5:} There exists a $L>0$, such that for $\vec{x}_v$, $\vec{x}^\prime_v$, $\vec{x}_u$, $\vec{x}_u^\prime$, and $(u,v)\in\mathcal{E}_v^h(t)$, we have
\begin{equation}
    ||\triangledown_{\vec{x}_v}g_{vu}(\vec{x}_v,\vec{x}_u)-\triangledown_{\vec{x}_v}g_{vu}(\vec{x}_v^\prime, \vec{x}_u)||\leq L||\vec{x}_v-\vec{x}_v^\prime||
\end{equation}

\begin{equation}
    ||\triangledown_{\vec{x}_v}g_{vu}(\vec{x}_v,\vec{x}_u)-\triangledown_{\vec{x}_v}g_{vu}(\vec{x}_v, \vec{x}^\prime_u)||\leq L||\vec{x}_u-\vec{x}_u^\prime||
\end{equation}

\textbf{Assumption 6:} The point $(\vec{x}^o,\vec{x}^o)\in\mathcal{X}^2$ exists such that for any $(u,v)\in\mathcal{E}_v^h(t)$, $|g_{vu}(\vec{x}_v^o,\vec{x}_u^o)|$ is bounded. Moreover, there exists a point $(\vec{x}_v^o,\vec{x}_u^o)\in\mathcal{X}^2$ such that for any $(v,u)\in\mathcal{N}_v^h(t)$, $||\triangledown_{\vec{x}_u}g_{vu}(\vec{x}_v^o,\vec{x}_u^o)||$ is bounded.   

\textbf{Proposition 1:} Following Assumptions 2, 4 and 5, the constraint function $g_{vu}(\cdot)$ for any $(v,u)\in\mathcal{N}_v^h(t)$ satisfies: (1) $|g_{vu}(\vec{x}_v,\vec{x}_u)-g_{vu}(\vec{x}_v^\prime,\vec{x}_u)|\leq B||\vec{x}_v-\vec{x}^\prime_v||$; $B>0$. (2) $|g_{vu}(\vec{x}_v,\vec{x}_u)|\leq C$; $C>0$.

\textbf{Proof:} The proof follows from the proof of Proposition 1 in \cite{yan2024decentralized}.

\textbf{Theorem:} The regret of any client $v$ running Algorithm 1 is upper bounded as
\begin{equation}
\begin{split}
Reg(T) = \mathbb{E}[\sum_{t=1}^T [f_{v,t}(\vec{x}_{v,t} - \sum_{t=1}^T f_{v,t}(\vec{x}_v)]\leq \\
\bigg(\frac{2r^2}{a^2}+mC^2 + VG^2\bigg)a\sqrt{T}+\\ \frac{5aG^2}{2}\sum_{v=1}^V \frac{1+\frac{1}{\zeta}\beta_{vu}}{1-(1+\zeta)\beta_{vu}}\sqrt{T} = \mathcal{O}(\sqrt{T})
\end{split}
\end{equation}
where $\eta =\frac{a}{\sqrt{T}}$, $a>0$ is a constant, $\zeta\in\bigg(0,\frac{1}{\beta}-1\bigg)$, $\delta = \frac{1}{4\eta^2}$, and $m = |\mathcal{E}|$.
Furthermore, for any $(v,u)\in\mathcal{E}_v^h(t)$, the constraint violation is upper bounded by
\begin{equation}
\begin{split}
 \mathbb{E}\bigg[\sum_{t=1}^T g_{vu}(\vec{x}_{v,t},\vec{x}_{u,t})\bigg] \leq 2\sqrt{VGr}\bigg(\frac{1}{a} + a\zeta + 2aB^2\bigg)^{1/2}T^{3/4}\\
 + (1+a^2\zeta + 2a^2B^2) \bigg(\frac{4r^2}{a^2}+ 2mC^2 +2VG^2\bigg)^{1/2} = \mathcal{O}(T^{3/4}).
 \end{split}
\end{equation}

\textbf{Proof:} The proof is omitted due to space constraints. However, it is a modification of the proof of Theorem 1 in \cite{yan2024decentralized}.

\section{Simulation Results}
We consider $45$ clients collaborating in a fully decentralized federated learning setting to train their local models. The graph network is fully connected. We assume there are $30$ Byzantine clients and $15$ honest clients, similar to \cite{yemini2025resilient}. The online federated learning setting is modeled as online decentralized logistic regression where the goal is to estimate the vector $\vec{x}$ by solving the following optimization problem among honest clients,
\begin{equation}
\begin{split}
  \min_{\vec{x}_{v,t}\in\mathcal{X},\forall v,t}\sum_{t=1}^T \sum_{i=1}^{V-b}\bigg[\log (1 + \exp(-l_{v,t}\psi_{v,t}^\mathbb{T}\vec{x}_{v,t})\bigg]  \\
  \text{subject to} \quad \frac{1}{T}\sum_{t=1}^T||\vec{x}_{v,t}-\vec{x}_{u,t}||^2 \leq \kappa_{vu}^2; \forall (v,u)\in\mathbb{E}_v^h(t)
  \end{split}
\end{equation}
where $D_{v,t}:=(l_{v,t},\psi_{v,t})$ is the data sample of client $v$ with label $l_{v,t}\in\{1,-1\}$ and feature vector $\psi_{v,t}\in\mathbb{R}^d$ is drawn from the Gaussian distribution $\mathcal{N}(\mathbf{0},\mathbf{I})$. The data sample is time-varying.  The step size $\eta = \frac{1}{\sqrt{T}}$, $T=1000$, and $||\vec{x}||\leq r=1$.

The trust values are generated as follows \cite{yemini2025resilient}: Let $\mathbb{E}[\alpha_{vu}(t)]=0.55$ if $u\in\mathcal{N}_v\cap \mathcal{V}_h$ and $\mathbb{E}[\alpha_{vu}(t)]=0.45$ if $u\in\mathcal{N}_v\cap\mathcal{V}_b$. The trust probability $\alpha_{vu}(t)$ is uniformly distributed in the interval $[\mathbb{E}[\alpha_{vu}(t)]-\frac{l}{2}, \mathbb{E}[\alpha_{vu}(t)+\frac{l}{2}]$, with $l$ chosen as $0.8$ with $|E_v|=|E_b|=0.05$, for every $v\in\mathcal{V}_h$ and $u\in\mathcal{V}_h\cup\mathcal{V}_b$. The honest clients are not aware of the values of $\mathbb{E}[\alpha_{vu}(t)]$ and $l$. The simulation results are averaged over $50$ system realizations due to the stochasticity of the trust values. We compare the time-average regret and constraint violation for the proposed algorithm with Byzantine clients and compare its performance with online Lagrangian descent algorithm for $15$ clients with no Byzantine attack.   

\begin{figure}[t!]
\centering
\includegraphics[width=2.5in]{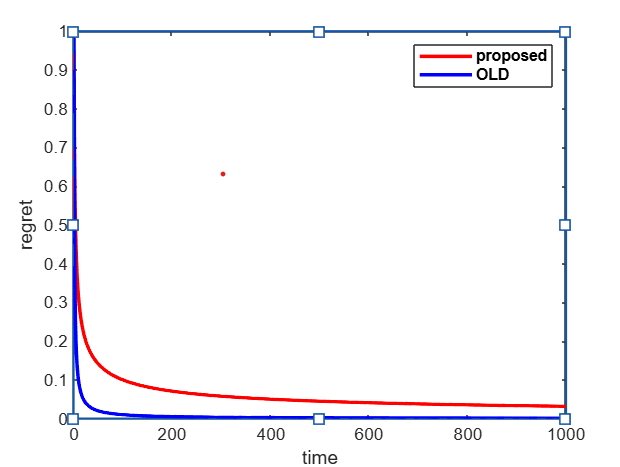}
\caption{Time-Average regret.}
\label{fig. 1}
\end{figure}

\begin{figure}[t!]
\centering
\includegraphics[width=2.5in]{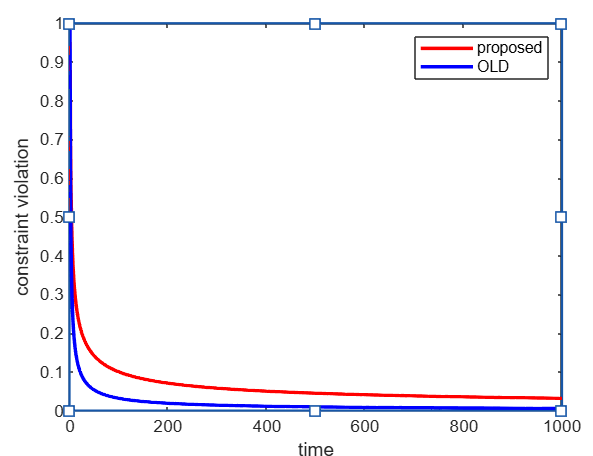}
\caption{Time-Average constraint violation.}
\label{fig. 1}
\end{figure}

Figures 1 and 2 show the time-average regret and constraint violation for the proposed algorithm with Byzantine attacks and online Lagrangian descent (OLD) with no Byzantine attack. It can be seen that the performance of the proposed algorithm is close to the Byzantine-free setting. It should be noted that there is no existing Byzantine-resilient approach that would incur sublinear regret when the number of Byzantine clients is higher than the number of honest clients. Hence, it is unnecessary to compare our proposed algorithms with other Byzantine-resilient approaches used in federated learning when Byzantine clients dominate the federation.

\section{Future Works}
In the future, we will theoretically find a bound that determines how much the number of Byzantine clients will exceed the number of honest clients before there is an eventual breakdown of the proposed algorithm. Moreover, we will determine the minimum iteration time and $T_f$ value sufficient to eliminate misclassification. We will also use various graph networks, such as the Erdos-Renyi graph and time-varying graphs, for the simulation.

\section{Conclusion}
This paper discussed how personalized models can be obtained in an online decentralized federated learning setting where clients communicate peer-to-peer and a majority of these clients are Byzantine. Currently, no Byzantine resilience approach used in federated learning can mitigate a dominating number of Byzantine clients. However, recent research in robotics shows that it is possible to leverage the cyber-physical properties of a system to predict malicious signals and thus mitigate attacks beyond data-based approaches commonly used in federated learning. Therefore, this paper addressed this challenge in federated learning for the first time by developing an online federated learning algorithm that can guarantee model personalization and trustworthiness by utilizing a trust probability obtained through the observation of cyber-physical systems. Through simulation, it is shown that the performance of the proposed algorithm in the presence of a dominating number of Byzantine clients is close to the performance of online Lagrangian descent, the benchmark algorithm used for the scenario where there are no Byzantine clients in the federation.  

\bibliographystyle{IEEEtran}
\bibliography{Ref.bib}
\end{document}